# 3D Surface Reconstruction of Underwater Objects

Prabhakar C J

Department of Studies in Computer Science
Kuvempu University, Shangaraghatta-577451
Karnataka, India

Praveen Kumar P U

Department of Studies in Computer Science
Kuvempu University, Shangaraghatta-577451
Karnataka, India

## ABSTRACT

In this paper, we propose a novel technique to reconstruct 3D surface of an underwater object using stereo images. Reconstructing the 3D surface of an underwater object is really a challenging task due to degraded quality of underwater images. There are various reason of quality degradation of underwater images i.e., non-uniform illumination of light on the surface of objects, scattering and absorption effects. Floating particles present in underwater produces Gaussian noise on the captured underwater images which degrades the quality of images. The degraded underwater images are preprocessed by applying homomorphic, wavelet denoising and anisotropic filtering sequentially. The uncalibrated rectification technique is applied to preprocessed images to rectify the left and right images. The rectified left and right image lies on a common plane. To find the correspondence points in a left and right images, we have applied dense stereo matching technique i.e., graph cut method. Finally, we estimate the depth of images using triangulation technique. The experimental result shows that the proposed method reconstruct 3D surface of underwater objects accurately using captured underwater stereo images.

## 1. INTRODUCTION

Optical imaging of the ocean floor offers scientists high levels of detail and ease of interpretation. However, light in underwater suffers from significant attenuation and backscatter, limiting the practical coverage of a single image to only a few square meters. For many scientific surveys, however, the area of interest is large, and can only be covered by hundreds or thousands of images acquired from a robotic vehicle or towed sled. Such surveys are required to study hydrothermal vents and spreading ridges in geology, ancient shipwrecks and settlements in archeology, forensic studies of modern shipwrecks and airplane accidents, and surveys of benthic ecosystems and species in biology.

In the field of 3D reconstruction, terrestrial applications have encouraged extensive work over the last three decades; on the other hand a limited amount of underwater applications have been explored primarily for mapping and positioning. Shape from X is a generic name for techniques that extract shape from images. Normally, in underwater environment the optical sensing techniques include Shape from Stereopsis [24], Shape from Photometric Stereo [15], Shape from Motion [7] and Active Stereo [14]. In the field of non-optic underwater sensing, acoustic cameras are employed for 3D mosaic reconstruction [2], whereas both acoustic and optic cameras are often used providing scene information that cannot be recovered from each sensor alone.

Three dimensional scene structures captured by a camera may be detected and acquired observing the apparent motion of brightness patterns from images. The primary visual motion cue useful for shape acquisition is the perceived movement of brightness patterns, known as optical flow [6] which is an approximation of the 3D world motion field. The 3D reconstruction from differential motion cues requires accurate optical flow computation. In the last fifteen years, theoretical developments in visual motion studies have established a unified framework for the treatment of the Structure from Motion (SFM) and Structure from Stereo (SFS) problems [3], also known as 3D Reconstruction from Multiple Views. 3D reconstruction from multiple views involves extracting target features from one image, matching and tracking these features across two or more images, and using triangulation to determine the position of the 3D target points relative to the camera. Visual motion methods have been well-studied, requiring densely-sampled image sequences. Instead, the trade-off in stereo vision is between the stereo correspondence problem and a more accurate and robust 3D reconstruction [19]. A large amount of works has addressed the correspondence problem, attempting to overcome the various difficulties of the large-displacement correspondence problem: occlusions, large rotations and disparities, photometric and projective distortions [12, 23].

The problem of outliers has been solved by the deployment of robust estimation methods [25]. The motion and structure may be estimate by using several recursive schemes. Extended Kalman Filter is the most popular approach for jointly estimation of motion and structure with satisfactory results. However, the computation cost of this approach grows cubically with the amount of features, causing a great bottleneck for real-time performance. Stereo matching is one of the most active research areas in computer vision. Stereo matching is a hard problem due to ambiguity in un-textured and occluded areas. Only dominant features, such as points of interest, can be matched reliably. This motivates the development of progressive approaches [19]. The reduced local disparity in search range makes progressive approaches very efficient in computation and robust. However, the seed initialization remains a computational bottleneck, although, robustness can be improved by enforcing the left-right symmetry constraint.

Recently, Graph Cuts and Belief Propagation furnish combinatorial optimization frameworks in which the performance for global stereo algorithms are considerably improved according to an evaluation framework applied to a standard reference stereo data set [19]. However, the underlying brightness constancy assumption of combinatorial optimization methods severely limits the range of their applications. The earliest attempts for 3D reconstruction use methods based on volume intersection, as Shape from Silhouette [10]. Traditional methods, such as stereo, handle large visibility changes between images by solving the correspondence problem between images. The most prominent approaches in 3-D reconstruction are the Voxel Coloring [22] and the Space Carving [9]. These approaches use the color consistency to distinguish surface points from other points in a scene. Cameras with an unoccluded view of a non-surface point see surfaces beyond the point, and hence inconsistent colors, in the direction of the point. Initially, the environment is represented as a discretized set of voxels, and then the algorithm is applied to color the voxels that are part of a surface in the scene. Another promising approach in 3D reconstruction is the Marching Cubes algorithm for rendering iso-surfaces from volumetric scan data [11]. The algorithm produces a triangle mesh surface representation by connecting the patches from all cubes on the iso-surface boundary.

In underwater scenario, research is directed at exploring the use of vision, potentially in conjunction of other sensors, to automatically control unmanned submersibles, including positioning and navigation by utilizing a photomosaic as a two-dimensional visual map. Recent activities combine video imagery taken from multiple views of a scene to derive size and depth measurements and 3D reconstructions. These activities support (semi-) autonomous or operator-supervised missions pertaining to automatic vision guided station keeping, location finding and navigation, survey and mapping, trajectory following and online reconstruction of a composite image, search and inspection of subsea structures. These tasks require an accurate estimation of camera position, together with fast, accurate correspondence determination, particularly for real-time registration. Common sources of error include non-planar seafloor, moving objects, illumination variations, transect superposition, positioning drift. A number of studies over the last several years have also addressed the 3D reconstruction for various applications. Khamene and Negahdaripour incorporate cues from stereo, motion and shading flow for 3-D reconstruction in underwater [8]; Majidi and Negahdaripour suggest the use of 3D reconstruction for global alignment of 3D sensor positions [13]; Nicosevici et al introduce 3D reconstruction from motion video and representation of the surface topography by piecewise planar surfaces for the construction of ortho-mosaics [16]. Hogue et al have developed a stereo vision-inertial sensing device deployed to reconstruct complex 3D structures in both the aquatic and terrestrial domains [5]. The sensor temporally combines 3D information, obtained using stereo vision algorithms with a 3 DOF inertial sensor. The resulting point cloud model is then converted to a volumetric representation and a textured polygonal mesh is extracted using the Marching Cubes algorithm [11].

In this paper, we have used stereo pair of cameras to capture the underwater images. We rectify captured underwater images using uncalibrated rectification technique. The dense stereo matching technique is applied to find correspondence points in both left and right images. We estimate the depth of an underwater image by using triangulation method. The dense depth map is converted to triangular surface meshes using the Delaunay triangulation algorithm. Thus, a texture mapping can be applied with relative easy and efficiency to the object.

The proposed method helps in the 3D surface reconstruction of underwater objects, which will be further helpful in recognition of underwater objects like mines, shipwrecks, pipelines and telecommunication cables, etc.

The organization of the paper is as follows: Section 2 presents the detailed description of steps involved in the proposed method like pre-processing of underwater images, uncalibrated stereo image rectification, stereo matching (correspondence) and depth estimation. Section 3 presents the experimental results. The Section 4 draws a conclusion.

## 2.3D RECONSTRUCTION

The proposed 3D surface reconstruction method for underwater objects involves several sequential steps. These steps are explained with detailed description of the algorithm.

### 2.1. Preprocessing

A major difficulty to process underwater images comes from light attenuation. Light attenuation limits the visibility distance, at about twenty meters in clear water and five meters or less in turbid water. The light attenuation process is caused by the absorption (which removes light energy) and scattering (which changes the direction of light path). Dealing with this difficulty, underwater imaging faces to many problems: first the rapid attenuation of light requires attaching a light source to the underwater vehicle providing the necessary lighting. Unfortunately, artificial lights tend to illuminate the scene in a non uniform fashion producing a bright spot in the center of the image and poorly illuminated area surrounding. Then, the floating particles highly variable in kind and concentration, increase absorption and scattering effects: they blur image features (forward scattering), modify colors and produce bright artifacts known as "marine snow".

We have adopted the preprocessing technique proposed by Stephene Bazielle et al. (2006) [1] to preprocess captured underwater images. This method involves sequence of preprocessing steps:

(1) Removing potential moir'e effect

(2) Resizing and extending symmetrically the image to get a squared image whose size is a power of two

(3) Converting color space from RGB to YCbCr

(4) Homomorphic filtering

(5) Wavelet denoising

(6) Anisotropic filtering

(7) Adjusting image intensity

(8) Converting from YCbCr to RGB and reverse symmetric extension

(9) Equalizing color mean

**(1) Removing potential moir'e effect.** A moir'e effect has the appearance of a wavy repetitive pattern on the image. It is not an underwater perturbation, and it is often considered as aliasing phenomena. Sampling moir'e mainly occurs in the analog to digital conversion process. Moir'e pattern is removed via spectral analysis by detecting peaks in the Fourier transform and deleting them assuming that they represent the moir'e effect [21]. Only few images suffer from moir'e degradation but removing it is important because the following processes enhance contrast so enhance the moir'e effect and consequently highly degrade results.

**(2) Resizing and extending symmetrically the image to get a squared image whose size is a power of two.** Symmetric extension prevents from potential border effects and resizing to squared image speeds up the following process by enabling to use fast Fourier transform and fast wavelet transform algorithms.

**(3) Converting color space from RGB to YCbCr (Luminance Chrominance).** This color space conversion allows us to work only on one channel instead of processing the three RGB channels. In YCbCr color space we process only the luminance channel (Y) corresponding to intensity component (gray scale image). The two other components correspond to chroma color difference. This step speeds up again all the following processings avoiding to process each time each RGB channels.

**(4) Homomorphic Filtering.** The homomorphic filtering is used to correct non uniform illumination and to enhance contrasts in the image. It's a frequency filtering, preferred to others techniques because it corrects non uniform lighting and sharpens the edges at the same time. We consider that image is a function of the product of the illumination and the reflectance as shown below.

$$f(x,y) = i(x,y) \cdot r(x,y), \qquad (1)$$

where $f(x,y)$ is the image sensed by the camera, $i(x,y)$ the illumination multiplicative factor, and $r(x,y)$ the reflectance function. If we take into account this model, we can assume that the illumination factor changes slowly through the view field, therefore it represents low frequencies in the Fourier transform of the image. On the contrary reflectance is associated with high frequency components. By multiplying these components by a high-pass filter we can then suppress the low frequencies i.e. the non uniform illumination in the image. The algorithm can be decomposed as follows:

—Separation of the illumination and reflectance components by taking the logarithm of the image. The logarithm converts the multiplicative effect into an additive one:

$$g(x,y) = \ln(f(x,y)) = \ln(i(x,y) \cdot r(x,y)) = \ln(i(x,y)) + \ln(r(x,y)). \ (2)$$

— Computation of the Fourier transform of the log-image

$$G(w_x, w_y) = I(w_x, w_y) + R(w_x, w_y). \qquad (3)$$

—The High-pass filter is applied to the Fourier transform which decreases the contribution of low frequencies (illumination) and also amplifies the contribution of mid and high frequencies (reflectance), sharpening the edges of the object in the image:

$$S(w_x, w_y) = H(w_x, w_y) \cdot I(w_x, w_y) + H(w_x, w_y) \cdot R(w_x, w_y), \qquad (4)$$

with, $H(w_x, w_y) = (r_H - r_L) \cdot \left( 1 - \exp\left( -\left( \frac{w_x^2 + w_y^2}{2\delta_w^2} \right) \right) \right) + r_L, \qquad (5)$

where $r_H = 2.5$ and $r_L = 0.5$ are the maximum and minimum coefficients values and $\delta_w$ is a factor which controls the cutoff frequency. These parameters are selected empirically.

—Computation of the inverse Fourier transform to come back in the spatial domain and then taking the exponent to obtain the filtered image.

**(5) Wavelet Denoising.** Homomorphic filter amplifies the noise present in the image to suppress it we go for denoising technique. Multi-resolution decompositions have shown significant advantages in image denoising. For this denoising filter we choose a nearly symmetric orthogonal wavelet bases with a bivariate shrinkage exploiting interscale dependency [20]. This wavelet denoising gives very good results compared to other denoising methods because, unlike other methods, it does not assume that the coefficients are independent. Indeed wavelet coefficients in natural image have significant dependencies. The algorithm can be decomposed as follows: we define the $(g)_+$ function as:

$$(g)_+ = \begin{cases} 0 & if \ \ g < 0 \\ g & otherwise \end{cases}$$

—Multiscale decomposition of the image corrupted by Gaussian noise using wavelet transform. We use Farras wavelet base (nearly symmetric filters for orthogonal 2-channel perfect reconstruction filter bank).

—Estimation of noise variance $\sigma_n^2$ using Eq. 6

$$\sigma_n^2 = median(|y_i|)/0.6745, \quad y_i \in subband \ HH \qquad (6)$$

—For each subband of each level except for the lowpass residual: Computation of the signal variance using (Eq. 7) and modification of the noisy wavelet coefficient according to the (Eq. 8).

$$\sigma = \sqrt{(\sigma_y^2 - \sigma_n^2)_+} , \qquad (7)$$

where $\sigma_n^2 = \frac{1}{M} \sum_{y_i \in N(k)} y_i^2$, $M$ the size of the neighborhood $N(k)$

$$y_i = \frac{\left( \sqrt{y_i^2 + y_{i+1}^2} - \frac{\sqrt{3}\sigma_n^2}{\sigma} \right)}{\sqrt{y_i^2 + y_{i+1}^2}} \cdot y_i , \qquad (8)$$

where $y_i$ is the child and $y_{i+1}$ its parent.

—Inversion of the multiscale decomposition to reconstruct the filtered image.

**(6) Anisotropic Filtering.** This filter smooths the image in homogeneous area, preserves edges and enhance them. It is used to smooth textures and reduce artifacts by deleting small edges amplified by homomorphic filtering. It also removes or attenuates unwanted artifacts and remaining noise. We follow the algorithm proposed by Perona and Malik [17]. This algorithm is automatic so it uses constant parameters selected manually. The previous step of denoising is very important to obtain good results with anisotropic filtering. Anisotropic filtering is usually used as long as result is not satisfactory. In our case we loop only few times set to constant value, to preserve a short computation time. One loop of the algorithm can be decomposed as follows: For each pixel

—Computation of the nearest-neighbor differences and computation of the diffusion coefficient in the four directions North, South, East, West. Many possibilities exist for this calculation; the easiest way is as follows:

$$\begin{aligned} \nabla_N I_{i,j} &= I_{i-1,j} - I_{i,j}, & c_{N_{i,j}} &= g(|\nabla_N I_{i,j}|) \\ \nabla_S I_{i,j} &= I_{i+1,j} - I_{i,j}, & c_{S_{i,j}} &= g(|\nabla_S I_{i,j}|) \\ \nabla_E I_{i,j} &= I_{i,j+1} - I_{i,j}, & c_{E_{i,j}} &= g(|\nabla_E I_{i,j}|) \\ \nabla_W I_{i,j} &= I_{i,j-1} - I_{i,j}, & c_{W_{i,j}} &= g(|\nabla_W I_{i,j}|) \end{aligned} \qquad (9)$$

where the function $g$ is defined as: $g(\nabla I) = e^{\left( -\left( \frac{\|\nabla I\|}{K} \right)^2 \right)}$ and with $K$ set to 0.1. This diffusion function favors high contrast edges over low contrast ones.

—Modification of the pixel value using below equation

$$I_{i,j} = I_{i,j} + \lambda[c_N \nabla_N I + c_S \nabla_S I + c_E \nabla_E I + c_W \nabla_W I]_{i,j}, \qquad (10)$$

with $0 \leq \lambda \leq 1/4$.

**(7) Adjusting image intensity.** This step increases contrast by adjusting image intensity values. It suppresses eventually outliers pixels to improve contrast stretching. It then stretches contrast to use the whole range of intensity channel and if necessary it saturates some low or high values.

**(8) Converting from YCbCr to RGB and reverse symmetric extension.** After this step luminance channel has been preprocessed, so to regain colors we convert back the image the RGB space, and cut out the symmetric extension part of the image to recover the image with original size.

**(9) Equalizing color mean.** In underwater imaging color channels are rarely balanced correctly. This step enables to suppress predominant color by equalizing RGB channels means. It is rather used to produce a more pleasant image than to better segmentation. Because segmentation is in general performed on gray image and color equalization does really not change the gray image.

## 2.2. Uncalibrated Stereo Image Rectification

Given a pair of stereo images, rectification determines a transformation of each image plane such that pairs of conjugate epipolar lines become collinear and parallel to one of the image axes. The important advantage of rectification is that computing stereo correspondences is reduced to a 1-D search problem along the horizontal raster lines of the rectified images. In the case of uncalibrated cameras, there are more degrees of freedom in choosing the rectifying transformation and a few competing methods are present in the literature. Each aims at producing a "good" rectification by minimizing a measure of distortion, but none is clearly superior to the others, not to mention the fact that there is no agreement on what the distortion criterion should be. In this paper we adopt Quasi-Euclidean epipolar rectification method for uncalibrated images proposed by Andrea Fussiello et al. (2010) [4].

Geometrically, in the Euclidean frame, rectification is achieved by a suitable rotation of both image planes. The correspondent image transformation is the collineation induced by the plane at infinity. As a result, the plane at infinity is the locus of zero-disparity in the rectified stereo pair. This is signified by saying that Euclidean rectification is done with respect to the plane at infinity. In the uncalibrated case the reference plane is generic, as any plane can play the role of the infinity plane in the projective space. Our uncalibrated rectification can be seen as referred to a plane that approximates the plane at infinity.

We assume that intrinsic parameters are unknown and that a number of corresponding points $m_l^j \leftrightarrow m_r^j$ are available. The method seeks the collineations that make the original points satisfy the epipolar geometry of a rectified image pair.

The fundamental matrix of a rectified pair has a very specific form, namely it is the skew-symmetric matrix associated with the cross-product by the vector $u_1 = (1,0,0)$:

$$[u_1]_\times = \begin{bmatrix} 0 & 0 & 0 \\ 0 & 0 & -1 \\ 0 & 1 & 0 \end{bmatrix}. \qquad (11)$$

Let $H_r$ and $H_l$ be the unknown rectifying collineations. When they are applied to the corresponding tie-points $m_l^j$, $m_r^j$ respectively, the transformed points must satisfy the epipolar geometry of a rectified pair, namely:

$$(H_r m_r^j)^T [u_1]_\times (H_l m_l^j) = 0. \qquad (12)$$

The left-hand side of (Eq. 12) is an algebraic error, i.e., it has no geometrical meaning, so we used instead the Sampson error that is a first order approximation of the geometric error. The matrix $F = H_r^T [u_1]_\times H_l$ can be considered as the fundamental matrix between the original images, therefore, in our case, the squared Sampson error for the $j^{th}$ correspondence is defined as:

$$E_j^2 = \frac{(m_r^{jT} F m_l^j)^2}{(F m_l^j)_1^2 + (F m_l^j)_2^2 + (m_r^{jT} F)_1^2 + (m_r^{jT} F)_2^2}, \qquad (13)$$

where $(\cdot)_i$ is the $i^{th}$ component of the normalized vector. As this equation must hold for any $j$, one obtains a system of non-linear equations $\{E_j = 0\}$ in the unknown $H_r$ and $H_l$. A least-squares solution can obtained with the Levenberg-Marquardt algorithm, but the way in which $H_r$ and $H_l$ are parameterized is crucial, and characterizes our approach with respect to the previous ones. We force the rectifying collineations to have the same structure as in the calibrated (Euclidean) case, i.e. to be collineations induced by the plane at infinity, namely

$$H_r = K_{nr} R_r K_{or}^{-1}, \qquad H_l = K_{nl} R_l K_{ol}^{-1}, \qquad (14)$$

The old intrinsic parameters $(K_{ol}, K_{or})$ and the rotation matrices $(R_l, R_r)$ are unknown, whereas the new intrinsic parameters $(K_{nl}, K_{nr})$ can be set arbitrarily, provided that vertical focal length and vertical coordinate of the principal point are same. Indeed, it is easy to verify that the matrix $K_{nr}^T [u_1]_\times K_{nl}$ is equal (up to scale) to $[u_1]_\times$, provided that the second and third row of $K_{nr}$ and $K_{nl}$ are the same. Hence it is not necessary to include the matrices $K_{nr}$ and $K_{nl}$ in the parameterization.

Each collineation depends in principle on five (intrinsic) plus three (rotation) unknown parameters. The rotation of one camera along its X-axis, however, can be eliminated. Consider the matrix

$$F = K_{or}^{-T} R_r^T [u_1]_\times R_l K_{ol}^{-1}. \qquad (15)$$

Let $R_r'$ and $R_l'$ be the same matrices as $R_r$ and $R_l$ after pre-multiplying with an arbitrary (but the same for both) rotation matrix about the X-axis. It is easy to verify that $R_r^T [u_1]_\times R_l = R_r'^T [u_1]_\times R_l'$. Geometrically, this coincide with rotating a rectified pair around the baseline, which do not alter the rectification, but, in a real camera, it affects the portion of the scene that is imaged. Accordingly, we set to zero the rotation around the X-axis of the left camera.

We further reduce the number of parameters by making as educated guess on the old intrinsic parameters: no skew, principal point in the center of the image, aspect ratio equal to one. The only remaining unknowns are the focal lengths of both cameras. Assuming that they are identical and equal to $\alpha$, we get:

$$K_{or} = K_{ol} = \begin{bmatrix} \alpha & 0 & w/2 \\ 0 & \alpha & h/2 \\ 0 & 0 & 1 \end{bmatrix}. \tag{16}$$

where $w$ and $h$ are width and height (in pixel) of the image. In summary, two collineations are parameterized by six unknowns: five angles and the focal length $\alpha$. Focal length is expected to vary in the interval $[1/3(w+h), 3(w+h)]$, so we consider instead the variable $\alpha' = \log_3(\alpha/(w+h))$ which varies in $[-1, 1]$.

The minimization of the cost function is carried out using Levenburg-Marquardt, starting with all the unknown variables set to zero. When $\alpha'$ converges outside the boundaries of the interval $[-1, 1]$ a random restart is attempted. If the problem persists the minimization is carried out with fixed $\alpha' = 0$.

Finally, the new intrinsic parameters ($K_{nr}$ and $K_{nl}$) are set equal to the old ones: $K_{nr} = K_{nl} = K_{ol}$, modulo a shift of the principal point, that might be necessary to center the rectified images in the customary image coordinate frame. Horizontal translation has no effect on the rectification, whereas vertical translation must be the same for both images.

## 2.3. Dense Stereo Matching

A large number of stereo matching algorithms exist and they can be classified into two main categories: local and global methods, according to the principle they are based upon. Other methods called cooperative algorithms use local and global approach at the same time. The difficulty is to choose an algorithm to perform a dense 3D reconstruction taking into account rendering, metrologic quality, computing speed and complexity of the scene. The taxonomy of [19] provides information about the overall performance of the principal algorithms (textureless regions, depth discontinuity regions, occluded regions). Finally, the graph-cut method gives excellent results, performing better in textureless areas and near discontinuities, and outperforming the other optimization methods. The major downsides are the relatively high computation time, and the need precisely tuned parameters, whose values are often image-dependent. This algorithm remains however a very interesting choice for our application, since the quality of the rendering process is a higher priority compared to execution time.

Roy S. and Cox I.J. [18] were the first ones to use this algorithm in the context of multi-camera stereovision. In order to explain the graph-cut method, we will concentrate on the case of graphs with only two terminals.

Figure 1 show a simple example of a two terminal graph, which can be used to minimize an energy function on a $3 \times 3$ image with two labels. The two terminals are usually called source s, and sink t. They correspond to the set of labels (different depths) that can be assigned to pixels of the image.

In the general case of graph-cut theory, the goal is to find a cut that has a minimum cost among all cuts, by minimizing an energy function.

Let function f be the disparity function associated to each pixel of an image. We search labeling f that minimizes the energy. To define this energy on a photo-consistence criterion (similarity between intensities of a pixel $\mathbf{p}$ in the first image and the pixel ($\mathbf{p} + fp$) in the second image) called data term. A second term, called smoothness term, penalizes discontinuities between neighborhood pixels. Thus, the energy can be written as:

$$E(f) = \sum_{p \in P} D_p(f_p) + \sum_{\{p,q\} \in N} V_{\{p,q\}}(f_p, f_q), \tag{17}$$

where term $D_p$ is the data term and $V(f_p, f_q)$ is the smoothness term penalty between adjacent pixels.

In (kolmogorov, zabih), the energy minimization considers the input images symmetrically, handles visibility properly, and imposes spatial smoothness while preserving discontinuities.

The disparity map gives the dense correspondence map between the stereo images. Thus, the depth map is computed by triangulation with matched point pairs and camera parameters. The mathematical equation for estimation of depth map is

$$z = \frac{f * B}{d}, \tag{18}$$

Where, z is the depth, f is focal length of the camera, B is Baseline between two cameras and d is the disparity.

The disparity map gives us a dense correspondence map between the stereo images. Thus, the depth map is computed by triangulation with matched point is considered independently. Therefore smoothing the surface is important to obtain a spatial coherence. In this paper, a spatial coherence is achieved by filtering the depth map.

## 3. EXPERIMENTAL RESULTS

To see the effectiveness of the proposed approach on underwater images, experiment is done in a small water body. The scene includes several objects at distance [1m, 2m], near the corner of the water body. The depth at the bottom of the water body was approximately 2m. The stereo setup consists of two identical digital still cameras, which are Canon D10 water proof. The two cameras are installed on a vertical stereo mount and it is kept in water body. The Figure 2 shows the captured underwater stereo images. The underwater images are preprocessed by applying sequentially homomorphic filtering, wavelet denoising and anisotropic filtering techniques. The Figure 3 shows the preprocessed underwater images.

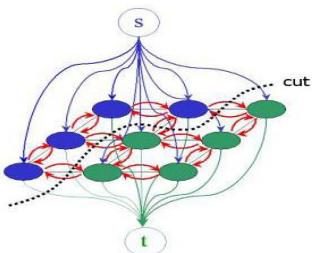

**Fig 1: Graph - Cut example**

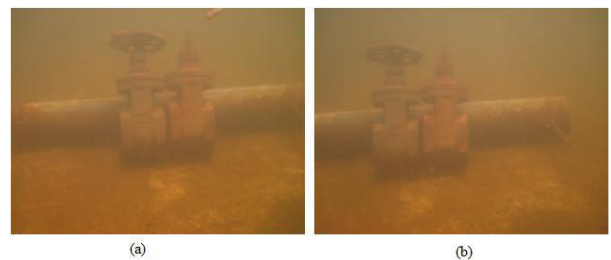

(a)  (b)

**Fig 2: a) Left Image and b) Right Image**

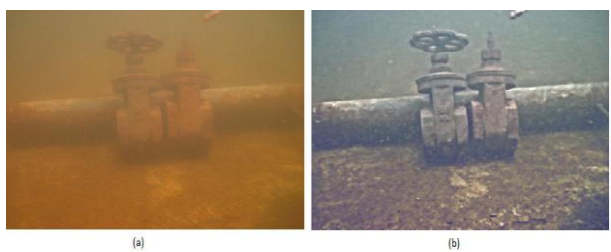

**Fig 3: a) Before Filtering and b) After Filtering**

The Figure 4 shows a result of uncalibrated rectification technique on left and right images. The Figure 5 shows the dense disparity map obtained with the implementation of the graph-cut method presented in (kolmogorov and zabih, 2002). Figure 6 shows the final reconstruction with texture mapping which can be used for 3D quantitative imaging. The dense depth map with texture mapping gives a 3D model with a good visual impression.

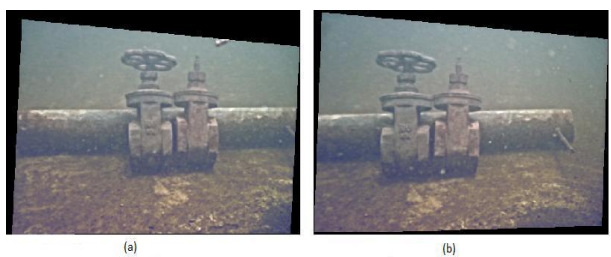

**Fig 4: a) Rectified Left Image and b) Rectified Right Image**

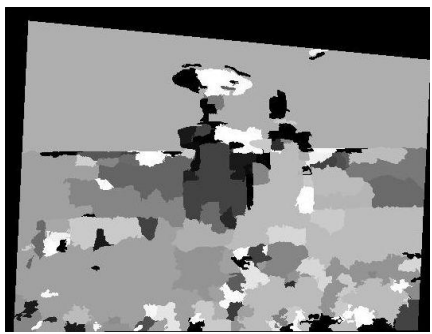

**Fig 5: Disparity Map**

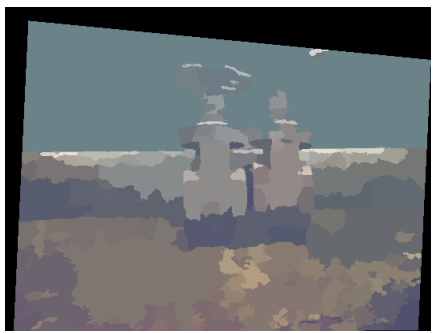

**Fig 6: Textured- Depth Map**

## 4. CONCLUSIONS

In this paper, we have presented a complete 3D surface reconstruction of small-scale underwater objects to supervised exploration of ocean floors. The first step concern with acquisition of stereo images by using stereo vision system. We preprocess the acquired stereo images sequentially by applying suitable filter. The preprocessed image is then rectified by using uncalibrated rectification technique. The rectified images are then used for stereo matching using Graph cut method. Finally, depth map is generated using triangulation technique. The experimental results shows that the proposed method reconstruct 3D surface of an underwater object accurately.

## 5. ACKNOWLEDGEMENTS

This work was supported by Naval Research Board (Grant No.158/SC/08-09), DRDO, New Delhi, India.

## 6. REFERENCES

[1] S. Bazeille I. Quidu, L. Jaulin, and J. P. Malkasse. Automatic underwater image pre-processing. In Proceedings of the Caracterisation du Milieu Marin (CMM'06), October 2006.

[2] U. Castellani, A. Fusiello, V. Murino, L. Papaleo, E. Puppo, S. Repetto, and M. Pittore. Efficient on-line mosaicing from 3d acoustical images. OCEANS'04 MTS/IEEE TECHNO OCEAN'04, 2:670–677, November 2004.

[3] Olivier D. Faugeras. What can be seen in three dimensions with an uncalibrated stereo rig? In Proceedings of the Second European Conference on Computer Vision, pages 563–578, May 1992.

[4] Andrea Fusiello and Luca Irsara. Quasi-euclidean epipolar rectification of uncalibrated images. Machine Vision and Applications, pages 1432–1769, May 2010.

[5] A. Hogue, A. German, and M. Jenkin. Underwater environment reconstruction using stereo and inertial data. In IEEE International Conference on Systems, Man and Cybernetics, pages 2372–2377, October 2007.

[6] B. Horn. Robot Vision. MIT Press, New York, 1986.

[7] A. Khamene, H. Madjdi, and S. Negahdaripour. 3-d mapping of sea floor scenes by stereo imaging. OCEANS'01 MTS/IEEE, 4:2577–2583, November 2001.

[8] Ali Khamene and Shahriar Negahdaripour. Motion and structure from multiple cues; image motion, shading flow, and stereo disparity. Computer Vision and Image Understanding, 90:99–127, April 2003.

[9] K. N. Kutulakos and Steven Seitz. What do n photographs tell us about 3d shape? In Technical Report TR680, Computer Science Dept., U. Rochester. January 1998.

[10] A. Laurentini. The visual hull concept for silhouette-based image understanding. IEEE Transactions on Pattern Analysis and Machine Intelligence, 16:150–162, February 1994.

[11] William E. Lorensen and Harvey E. Cline. Marching cubes: A high resolution 3d surface construction algorithm. ACM SIGGRAPH Computer Graphics, 21:163–169, July 1987.

[12] Bruce D. Lucas and Takeo Kanade. An iterative image registration technique with an application to stereo vision. In Proceedings of the 7th international joint conference on Artificial intelligence, volume 2, 1981.

[13] H. Madjidi and S. Negahdaripour. Global alignment of sensor positions with noisy motion measurements. IEEE Transactions on Robotics and Automation, 21:1092–1104, December 2005.

[14] S.G. Narasimhan, S.K. Nayar, B Sun, and S.J. Koppal. Structured light in scattering media. Proceedings of the Tenth IEEE International Conference on Computer Vision, 1:420–427, October 2005.

[15] S. Negahdaripour, H. Zhang, and X. Han. Investigation of photometric stereo method for 3-d shape recovery from underwater imagery. OCEANS'02 MTS/IEEE, 2:1010–1017, October 2002.

[16] T. Nicosevici, S. Negahdaripour, and R. Garcia. Monocular based 3-d seafloor reconstruction and ortho-mosaicing by piecewise planar representation. In Proceedings of MTS/IEEE OCEANS, volume 2, 2005.

[17] P. Perona and J. Malik. Scale-space and edge detection using anisotropic diffusion. IEEE Transactions on Pattern Analysis and Machine Intelligence, 12:629–639, July 1990.

[18] Sebastien Roy and Ingemar J. Cox. A maximum-flow formulation of the n-camera stereo correspondence problem. In Proceedings of the Sixth International Conference on Computer Vision, pages 492–499, January 1998.

[19] Daniel Scharstein and Richard Szeliski. A taxonomy and evaluation of dense two-frame stereo correspondence algorithms. International Journal of Computer Vision, 47:7–42, December 2002.

[20] L. Sendur and I. W. Selesnick. Bivariate shrinkage functions for wavelet-based denoising exploiting interscale dependency. IEEE Transactions on Signal Processing, 50:2744–2756, November 2002.

[21] D. N. Sidorov and ANil C. Kokaram. Suppression of moir'e patterns via spectral analysis. In Proceedings of SPIE in Visual Communications and Image Processing, January 2002.

[22] Greg Slabaugh, Bruce Culbertson, Tom Malzbender, and Ron Schafer. A survey of methods for volumetric scene reconstruction from photographs. In Proceedings International Workshop on Volume Graphics, pages 81–100, June 2001.

[23] Carlo Tomasi and Takeo Kanade. Shape and motion from image streams: a factorization method - part 3 detection and tracking of point features. Technical Report CMU-CS-91-132, Computer Science Department, Pittsburgh, PA, April 1991.

[24] H. Zhang. Automatic sensor platform positioning and 3-d target modeling from underwater stereo sequences. PhD Thesis, December 2005.

[25] Zhengyou Zhang, Rachid Deriche, Olivier Faugeras, and Quang-Tuan Luong. A robust technique for matching two uncalibrated images through the recovery of the unknown epipolar geometry. Artificial Intelligence, 78:87–119, October 1995.